\documentclass[9pt,technote,compsoc]{IEEEtran}

\usepackage{ifpdf}

\ifCLASSOPTIONcompsoc
  \usepackage[nocompress]{./sty/cite}
\else
  \usepackage{cite}
\fi

\usepackage[dvipdfmx]{graphicx}
\usepackage[dvipdfmx]{color}

\usepackage{amsmath}
\usepackage{amssymb}
\usepackage{array}

\ifCLASSOPTIONcompsoc
 \usepackage[caption=false,font=footnotesize,labelfont=sf,textfont=sf]{subfig}
\else
  \usepackage[caption=false,font=footnotesize]{subfig}
\fi

\usepackage{url}

\usepackage{slashbox}
\usepackage{booktabs}
\usepackage{multirow}
\usepackage{comment}
\usepackage{amsopn}
\usepackage[export]{adjustbox} %subfigで頭の位置を上に揃えるために使用

\graphicspath{{fig/}}
\newcommand{\Tref}[1]{Table~\ref{#1}}
\newcommand{\Eref}[1]{Eq.~\ref{#1}}
\newcommand{\Fref}[1]{Fig.~\ref{#1}}

\newcommand{\Sref}[1]{Section \ref{#1}}

\begin{document}

\title{Significance of Softmax-based Features\\in Comparison to\\Distance Metric Learning-based Features}

\author{Shota~Horiguchi,~\IEEEmembership{Member,~IEEE},
Daiki~Ikami,~\IEEEmembership{Student member,~IEEE} and~Kiyoharu~Aizawa,~\IEEEmembership{Fellow,~IEEE}% <-this % stops a space
\IEEEcompsocitemizethanks{\IEEEcompsocthanksitem S.Horiguchi was and D.Ikami, K.Aizawa are with the Department
of Information and Communication Engineering, The University of Tokyo, Bunkyo, Tokyo, Japan
133-8656.\protect\\
E-mail: see http://www.hal.t.u-tokyo.ac.jp/}
%\IEEEcompsocthanksitem J. Doe and J. Doe are with Anonymous University.}% <-this % stops an unwanted space
}


\markboth{Journal of \LaTeX\ Class Files, April~2019}%
{Shell \MakeLowercase{\textit{et al.}}: Bare Demo of IEEEtran.cls for Computer Society Journals}


\IEEEtitleabstractindextext{%
\begin{abstract}
End-to-end distance metric learning (DML) has been applied to obtain features useful in many computer vision tasks. However, these DML studies have not provided equitable comparisons between features extracted from DML-based networks and softmax-based networks. In this paper, we present objective comparisons between these two approaches under the same network architecture. 
\end{abstract}

}

\maketitle

\IEEEdisplaynontitleabstractindextext

% \fi
\IEEEpeerreviewmaketitle

\vspace{1cm}

\IEEEraisesectionheading{\section{Introduction}\label{sec:introduction}}

Recent developments in deep convolutional neural networks have made it possible to classify many classes of images with high accuracy.
It has also been shown that such classification networks work well as feature extractors.
Features extracted from classification networks show excellent performance in image classification \cite{donahue2014decaf}, detection, and retrieval \cite{razavian2014cnn,liu2015deepindex}, even when they have been trained to classify 1000 classes of the ImageNet dataset \cite{russakovsky2015imagenet}.
It has also been shown that fine-tuning for target domains further improves the features' performance \cite{wan2014deep,babenko2014neural}.

\begin{figure*}[h!]
	\centering
	\hfill
	\subfloat[Siamese ($dim=2$)]{
		\includegraphics[height=0.22\linewidth]{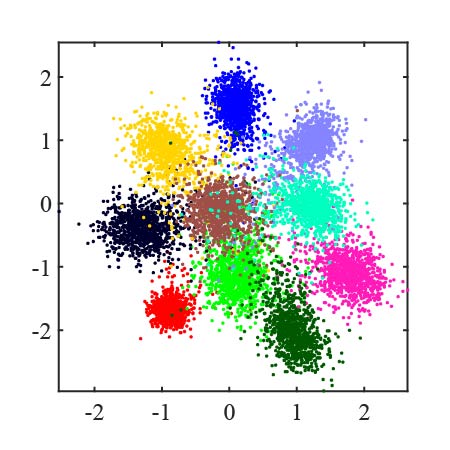}
		\label{fig1:siamese}
	}
	\hfill
	\subfloat[Softmax ($dim=2$)]{
		\includegraphics[height=0.22\linewidth]{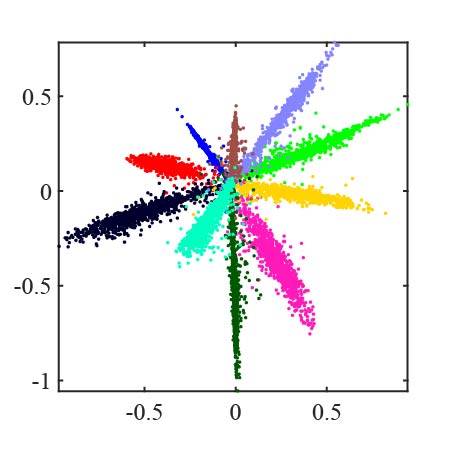}
		\label{fig1:softmax2}
	}\hfill
	\subfloat[Softmax ($dim=3$) + L2 normalization]{
		\includegraphics[height=0.22\linewidth]{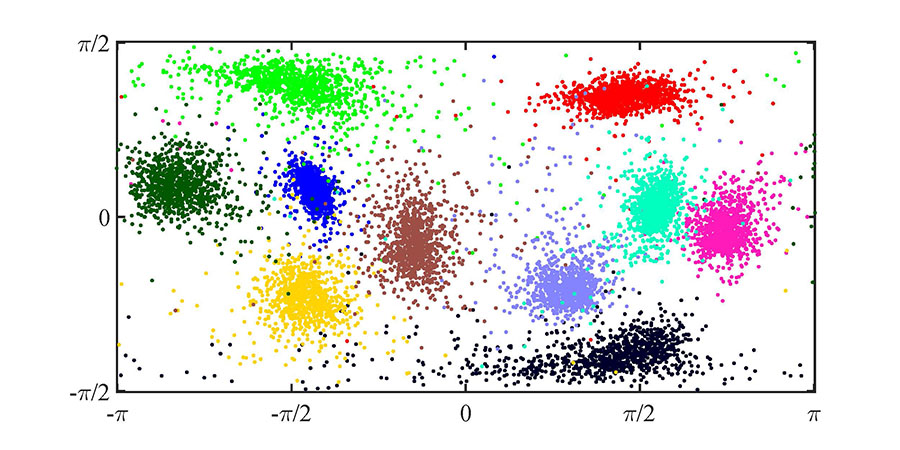}
		\label{fig1:softmax3}
	}
	\hfill
	\subfloat{
		\raisebox{2.5mm}{\includegraphics[height=0.18\linewidth, bb=0 0 83 496]{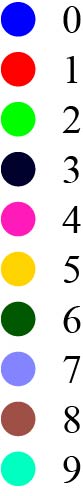}}
		\label{c}
	}
	\caption{Depiction of MNIST dataset. (a) Two-dimensional features obtained by siamese network. (b) Two-dimensional features extracted from softmax-based classifier; these features are well separated by angle but not by Euclidean norm. (c) Three-dimensional features extracted from softmax-based classifier; we normalized these to have unit L2 norm and depict them in an azimuth--elevation coordinate system. The three-dimensional features are well separated by their classes.}
	\label{fig:fig1}
\end{figure*}

On the other hand, distance metric learning (DML) approaches have recently attracted considerable attention. 
These obtain a feature space in which distance corresponds to class similarity; it is not a byproduct of the classification network.
End-to-end distance metric learning is a typical approach to constructing a feature extractor using convolutional neural networks and has been the focus of numerous studies \cite{bell2015productnet,schroff2015facenet,song2016deep,sohn2016improved,song2017learnable}.

However, there have been no experiments comparing softmax-based features with DML-based features under the same network architecture or with adequate fine-tuning.
An analysis providing a true comparison of DML features and softmax-based features is long overdue.

Figure \ref{fig:fig1} depicts the feature vectors extracted from a softmax-based classification network and a metric learning-based network.
We used LeNet architecture for both networks, and trained on the MNIST dataset \cite{lecun1998gradient}.
For DML, we used the contrastive loss function \cite{hadsell2006dimensionality} to map images in two-dimensional space.
For softmax-based classification, we added a two- or three-dimensional fully connected layer before the output layer for visualization.
DML succeeds in learning feature embedding (\Fref{fig1:siamese}).
Softmax-based classification networks can also achieve a result very similar to that obtained by DML--- Images are located near one another if they belong to the same class and far apart otherwise (Figures \ref{fig1:softmax2} and \ref{fig1:softmax3}).

Our contributions in this paper are as follows:
\begin{itemize}
	\item We show methods to exploit the ability of deep features extracted from softmax-based networks, such as normalization and proper dimensionality reduction. They are technically not novel, but they must be used for fair comparison between the image representations.
	\item We demonstrate that deep features extracted from softmax-based classification networks show competitive, or better results on clustering and retrieval tasks comparing to those from state-of-the-art DML-based networks \cite{song2016deep,sohn2016improved,song2017learnable} on the Caltech UCSD Birds 200-2011 dataset and the Stanford Cars 196 dataset.
	\item We show how the clustering and retrieval performances of softmax-based features and DML features change according to the size of the dataset. DML features show competitive or better performance in the Stanford Online Product dataset which consists of very small number of samples per class.
	\item We show that L2 normalization of softmax-based features is a powerful way to improve their performance. Even though we introduce probability invariant shift, which removes effects of softmax ambiguity and null space ambiguity, L2 normalization still works better.
\end{itemize}

In order to align the condition of the network architecture, we restrict the network architecture to GoogLeNet \cite{szegedy2015going} which has been used in state-of-the-art of DML studies \cite{song2016deep,sohn2016improved,song2017learnable}.

%----------------------------------------------------------------
\section{Background}
\subsection{Previous Work}
\subsubsection{Softmax-Based Classification and Repurposing of the Classifier as a Feature Extractor}
Convolutional neural networks have demonstrated great potential for highly accurate image recognition \cite{krizhevsky2012imagenet,simonyan2015very,szegedy2015going,he2016deep}.
It has been shown that features extracted from classification networks can be repurposed as a good feature representation for novel tasks \cite{donahue2014decaf,razavian2014cnn,qian2015fine} even if the network was trained on ImageNet \cite{russakovsky2015imagenet}.
For obtaining better feature representations, fine-tuning is also effective \cite{babenko2014neural}.

\subsubsection{Deep Distance Metric Learning}
Distance metric learning (DML), which learns a distance metric, has been widely studied 
\cite{bromley1994signature, chopra2005learning,chechik2010large, qian2015fine}.
Recent studies have focused on end-to-end deep distance metric learning \cite{bell2015productnet,schroff2015facenet,hoffer2015deep,song2016deep,
sohn2016improved,song2017learnable }.
However, in most studies comparisons of end-to-end DML with features extracted from classification networks have not been performed using architectures and conditions suited to enable a true comparison of performance.

Bell and Bala\cite{bell2015productnet} compared classification networks and siamese networks, but they used coarse class labels for classification networks and fine labels for siamese networks; thus, it was left unclear whether siamese networks are better for feature-embedding learning than classification networks.
Schroff et al.\cite{schroff2015facenet} used triplet loss for deep metric learning in their FaceNet, which showed performance that was state-of-the-art at the time, but their network was deeper than that of the previous method (Taigman et al.\cite{taigman2014deepface}); thus, triplet loss might not have been the only reason for the performance improvement, and the contribution from adopting triplet loss remains uncertain. Song et al.\cite{song2016deep} used lifted structured feature embedding; however, they only compared their method with a softmax-based classification network pretrained on ImageNet (Russakovsky et al.,\cite{russakovsky2015imagenet}) and did not compare it with a fine-tuned network.  Sohn\cite{sohn2016improved}, and Song et al.\cite{song2017learnable} also compared their methods to lifted structured feature embedding, thus the comparisons with softmax-based features have not been shown.

%-------------------------------------------------------------------------
\subsection{Differences Between Softmax-based Classification and Metric Learning}
\begin{figure}[t]
	\centering
	\includegraphics[width=\linewidth, bb=0 0 1200 460]{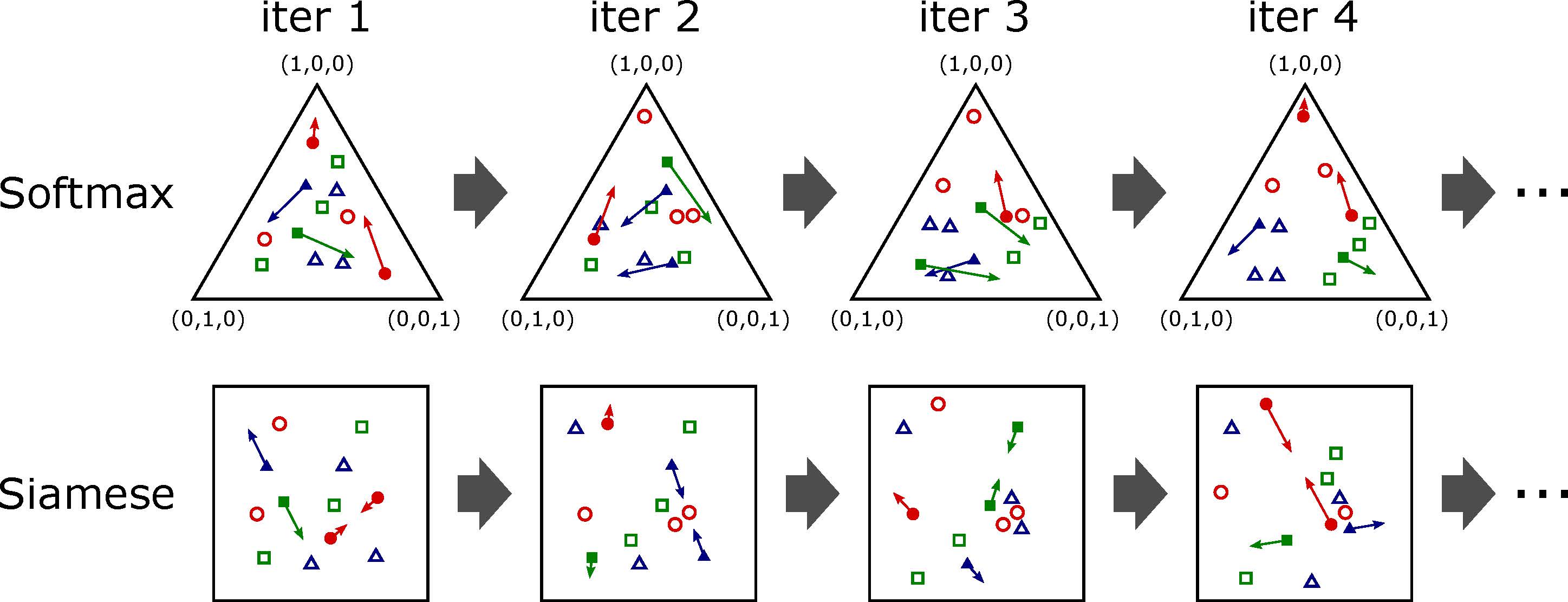}
	\caption{Illustration of learning processes for softmax-based classification network and siamese-based DML network. For softmax, the gradient is defined by the distance between a sample and a fixed one-hot vector; for siamese by the distance between samples.}
	\label{fig:softmaxAndDML}
\end{figure}

For classification, the softmax function (\Eref{eq:softmax}) is typically used:
\begin{equation}
\label{eq:softmax}
p_c=\frac{\exp(u_c)}{\sum_{i=1}^{C} \exp(u_i)},
\end{equation}
where $p_c$ denotes the probability that the vector $\mathbf{u}$ belongs to the class $c$.
The loss of the softmax function is defined by the cross-entropy
\begin{equation}
E=-\sum_{c=1}^{C} q_c\log p_c,
\end{equation}
where $\mathbf{q}$ is a one-hot encoding of the correct class of $\mathbf{u}$.
To minimize the cross-entropy loss, networks are trained to make the output vector $\mathbf{u}$ close to its corresponding one-hot vector.
It is important to note that the target vectors (the correct outputs of the network) are fixed during the entire training (\Fref{fig:softmaxAndDML}).

On the other hand, DML methods use distance between samples.
They do not use the values of the labels; rather, they ascertain whether the labels are the same between target samples.
For example, contrastive loss \cite{hadsell2006dimensionality} considers the distance between a pair of samples. 
Recent studies \cite{schroff2015facenet,song2016deep,song2017learnable,sohn2016improved} use pairwise distances between three or more images at the same time for fast convergence and efficient calculation.
However, these methods have some drawbacks.
For DML, in contrast to optimization of the softmax cross-entropy loss, the optimization targets are not always consistent during training even if all possible distances within the mini-batch are considered.
Thus, the DML optimization converges slowly and is not stable.

\begin{figure}[t]
	\centering
	\subfloat[GoogLeNet (dimensionality is reduced to n by PCA)]{
		\includegraphics[width=0.9\linewidth, bb=0 0 2933 617]{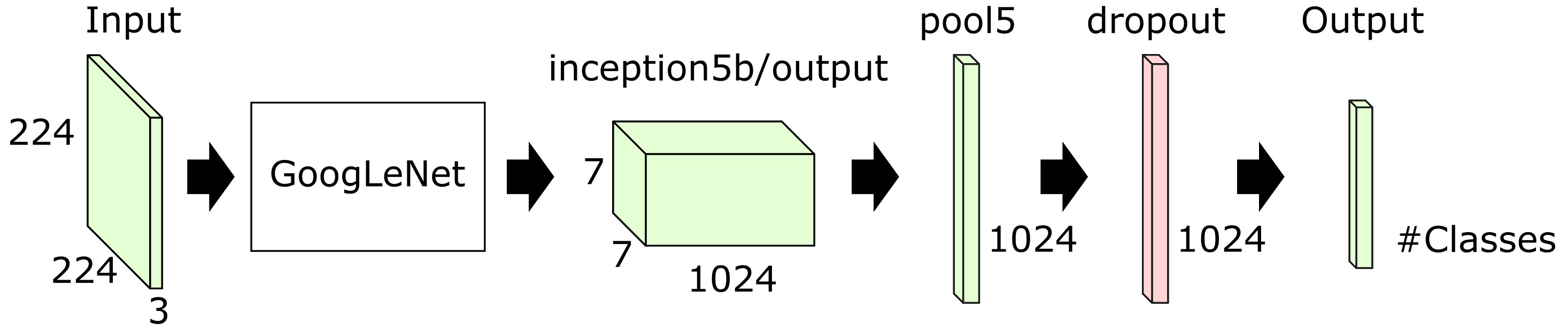}
		\label{fig:googlenet}
	}
	\\
	\subfloat[GoogLeNet with dimensionality reduction by a fully connected layer just before the output layer (FCR1)]{
		\includegraphics[width=\linewidth, bb=0 0 3400 617]{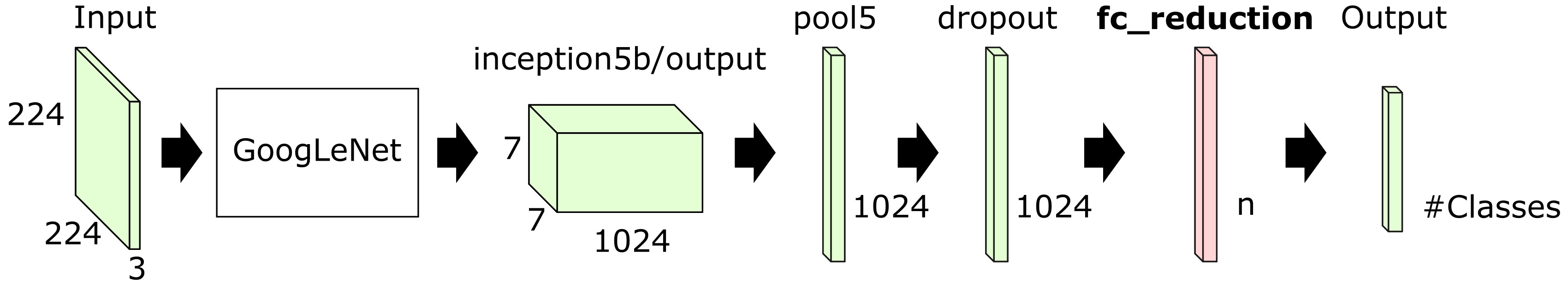}
		\label{fig:googlenet_fc1}
	}
	\\
	\subfloat[GoogLeNet with dimensionality reduction by a fully connected layer followed by a dropout layer (FCR2)]{
		\includegraphics[width=\linewidth, bb=0 0 3400 617]{./fig/dimension/googlenet_fc1.jpg}
		\label{fig:googlenet_fc2}
	}
	\caption{	GoogLeNet \cite{szegedy2015going} architecture we use in this paper. We extracted the features of the red-colored layers. For (a), we applied PCA to reduce the number of feature dimensions. For (b) and (c), the dimensionality is reduced by the fc\_reduction layer.}
\end{figure}

%\begin{table}[tbp]
\begin{table}[h]
	\centering
	\caption{Properties of datasets used in our experiments. Each cell shows the number of images (upper figure) and the number of classes (lower figure).}
	\label{tbl:clustering_dataset}
	\begin{tabular}{cccc}\toprule
		Dataset&Train&Test&Total\\\midrule
		\multirow{2}{*}{CUB \cite{wah2011caltech}}&5,864&5,924&11,788\\
		&100&100&200\\\midrule
		\multirow{2}{*}{CAR \cite{krause20133d}}&8,054&8,131&16,185\\
		&98&98&196\\\midrule
		\multirow{2}{*}{OP \cite{song2016deep}}&59,551&60,502&120,053\\
		&11,318&11,316&22,634\\\bottomrule
	\end{tabular}
\end{table}

\begin{table*}
	\centering
	\caption{CUB: NMI (clustering) and Recall@K (retrieval) scores for the test set of the Caltech UCSD Birds 200-2011 (CUB) dataset.}
	\label{tbl:CUB}
	\begin{tabular}{lcccccc}\toprule
		&&(clustering)&\multicolumn{4}{c}{Recall@K (retrieval)}\\\cmidrule(lr){4-7}
		&dim&NMI&K=1&K=2&K=4&K=8\\\midrule
		Lifted struct \cite{song2016deep}&64&56.5&43.6&56.6&68.6&79.6\\
                                                          &64&(56.0)&(42.7)&(55.0)&(67.2)&(78.1)\\
		N-pair loss \cite{sohn2016improved}&64&57.2&45.4&58.4&69.5&79.5\\
		Clustering loss \cite{song2017learnable}&64&59.2&48.2&61.4&71.8&81.9\\
		Random projection + L2&64&56.9&47.5&60.1&71.9&81.6\\
		PCA + L2&64&\textbf{60.8}&\textbf{51.1}&\textbf{64.0}&\textbf{75.3}&\textbf{84.0}\\
		FCR1 + L2&64&59.1&49.0&61.1&72.7&82.3\\
		FCR2 + L2&64&57.4&48.0&60.3&72.2&81.6\\\bottomrule
	\end{tabular}
\end{table*}
\begin{table*}
	\centering
	\caption{CAR: NMI (clustering) and Recall@K (retrieval) scores for the test set of the Stanford Cars 196 (CAR) dataset.}
	\label{tbl:CAR}
	\begin{tabular}{lcccccc}\toprule
		&&(clustering)&\multicolumn{4}{c}{Recall@K (retrieval)}\\\cmidrule(lr){4-7}
		&dim&NMI&K=1&K=2&K=4&K=8\\\midrule
		Lifted struct \cite{song2016deep}&64&56.9&53.0&65.7&76.0&84.0\\
		                                             &64&(57.1)&(50.5)&(63.6)&(74.9)&(83.6)\\
		N-pair loss \cite{sohn2016improved}&64&57.8&53.9&66.8&77.8&86.4\\
		Clustering loss \cite{song2017learnable}&64&59.0&58.1&70.6&80.3&87.8\\
		Random projection + L2&64&53.6&63.5&74.4&83.2&89.6\\
		PCA + L2&64&58.3&\textbf{69.4}&\textbf{80.0}&\textbf{87.2}&\textbf{92.4}\\
		FCR1 + L2&64&58.7&66.7&77.7&85.2&90.8\\
		FCR2 + L2&64&\textbf{60.4}&67.9&78.4&86.1&91.3\\\bottomrule
	\end{tabular}
\end{table*}
\begin{table*}
	\centering
	\caption{OP: NMI (clustering) and Recall@K (retrieval) scores for the test set of the Online Product (OP) dataset.}
	\label{tbl:OP}
	\begin{tabular}{lccccc}\toprule
		&&(clustering)&\multicolumn{3}{c}{Recall@K (retrieval)}\\\cmidrule(lr){4-6}
		&dim&NMI&K=1&K=10&K=100\\\midrule
		Lifted struct \cite{song2016deep}&64&88.7&62.5&80.8&91.9\\
		                                             &64&(87.7)&(61.0)&(79.9)&(91.5)\\
		N-pair loss \cite{sohn2016improved}&64&89.4&66.4&83.2&93.0\\
		Clustering loss \cite{song2017learnable}&64&\textbf{89.5}&\textbf{67.0}&\textbf{83.7}&\textbf{93.2}\\
		Random projection + L2&64&85.5&54.3&69.6&81.4\\
		PCA + L2&64&87.5&62.4&78.9&89.7\\
		FCR1 + L2&64&87.7&61.3&78.6&90.1\\
		FCR2 + L2&64&87.9&62.5&79.8&90.8\\\bottomrule
	\end{tabular}
\end{table*}

%-------------------------------------------------------------------------
\section{Methods}

\subsection{Dimensionality Reduction Layer}
\label{sec:dimensionaliry_reduction_layer}

One of DML's strength in using fine-tuning is the flexibility of its output dimensionality by a final fully connected layer.
When using features of a mid-layer of a softmax classification network, on the other hand, the dimensionality of the features is fixed.
Some existing methods \cite{babenko2014neural} use PCA or discriminative dimensionality reduction to reduce the number of feature dimensions.
In our experiment, we evaluated three methods for changing the feature dimensionality.
Following conventional PCA approaches, we extracted features from a 1024-dimensional pool5 layer of GoogLeNet \cite{szegedy2015going} (\Fref{fig:googlenet}) and applied PCA to reduce the dimensionality.
As a comparison, we also tried random projection for dimensionality reduction via orthogonal projection matrix.
In a contrasting approach, we made use of a fully connected layer---we added a fully connected layer having the required number of neurons just before the output layer (FCR1, \Fref{fig:googlenet_fc1}).
We also investigated a third approach in which a fully connected layer is added followed by a dropout layer (FCR2, \Fref{fig:googlenet_fc2}).

\subsection{Normalization}
In this study, all the features extracted from the classification networks are from the last layer before the last output layer.
The outputs are normalized by the softmax function and then evaluated by the cross-entropy loss function in the networks.
The output vector $\mathbf{p}=\left(p_i\right)$ is given by $\mathrm{softmax}\left(\mathbf{y}\right)$.
For an arbitrary constant $c$, $\mathrm{softmax}\left(\mathbf{y}\right)$ equals to $\mathrm{softmax}\left(\mathbf{y}+c\mathbf{1}\right)$.
The features $\mathbf{x}$ we extracted from the networks are given as $\mathbf{y}=W\mathbf{x}+\mathbf{b}$, where $W$ and $\mathbf{b}$ are from the linear projection matrix and the bias, respectively.
As pointed out, the vector $\mathbf{y}$ has an ambiguity in the softmax function, thus $\mathbf{x}$ should be normalized for the use of deep features.

In this paper, we show that L2 normalization is empirically effective.
Some studies used L2 normalization for deep features extracted from softmax-based classification networks \cite{taigman2014deepface,babenko2014neural}, whereas many recent studies have used the features without any normalization \cite{krizhevsky2012imagenet,song2016deep,wei2016dense}.
$W\mathbf{x}$ and $W\mathbf{x}/|\mathbf{x}|$ do not always result in the same probabilities after the softmax function is applied.
Applying L2-normalization for deep features rounds the confidence of predicted results while it keeps the magnitude relationship between probabilities of every classes.
However, as \Fref{fig1:softmax2} clearly indicates, the distance between features extracted from a softmax-based classifier should be evaluated by cosine similarity, not by the Euclidean distance.
In this study, we mainly validated the efficiency of L2 normalization of deep features.

We also considered another way to cope with the ambiguity introduced by the shift invariance of softmax function and null space of $W$.
We define a distance metric that takes softmax invariance and the null space into account, which treats features that result in the same probabilities as equal.
We report the experimental results of using the distance metric with probability invariant shift in \Sref{ssec:shift_vs_l2}.

\begin{figure*}[t!]
	\begin{minipage}{0.90\linewidth}
\hfill
		\begin{minipage}{0.45\hsize}
			\centering
			\includegraphics[width=\linewidth, bb=0 0 600 453]{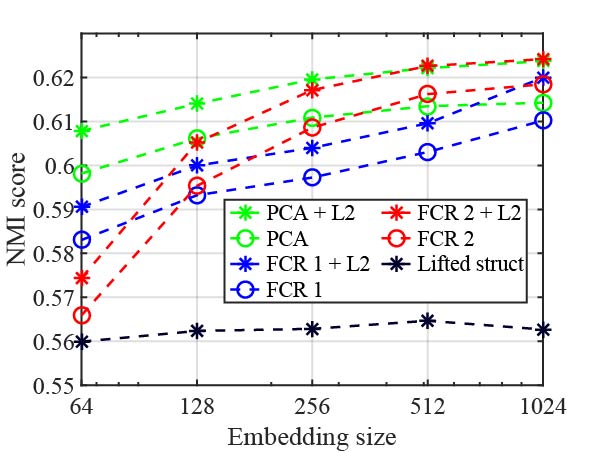}
		\end{minipage}
		\hfill
		\begin{minipage}{0.45\hsize}
			\centering
			\includegraphics[width=\linewidth, bb=0 0 600 453]{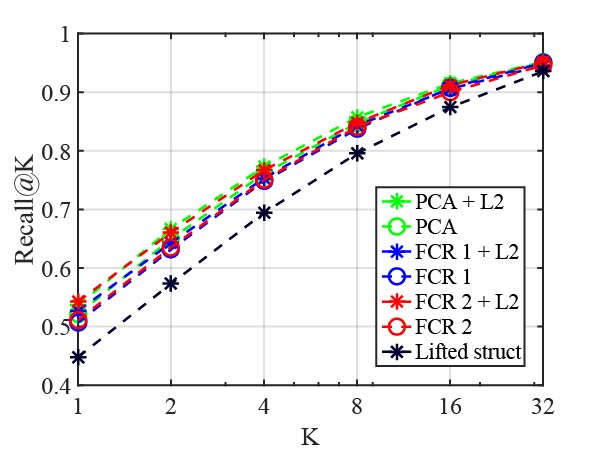}
		\end{minipage}	
		\caption{CUB: Comparisons between softmax-based features and lifted structured feature embedding \cite{song2016deep} on NMI (clustering), and Recall@K (retrieval) scores for the test set of the Caltech UCSD Birds 200-2011 (CUB) dataset. The dimension of the feature used in the retrieval experiments is 64.}
		\label{fig:CUB}
	\end{minipage}
	\\
	\begin{minipage}{0.90\linewidth}
\hfill
		\begin{minipage}{0.45\hsize}
			\centering
			\includegraphics[width=\linewidth, bb=0 0 600 453]{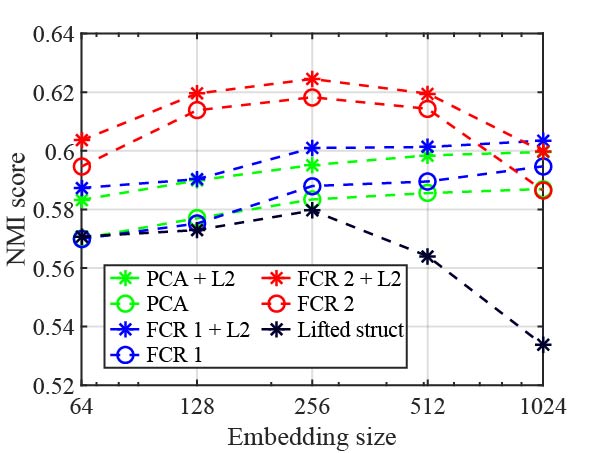}
		\end{minipage}
		\hfill
		\begin{minipage}{0.45\hsize}
			\centering
			\includegraphics[width=\linewidth, bb=0 0 600 453]{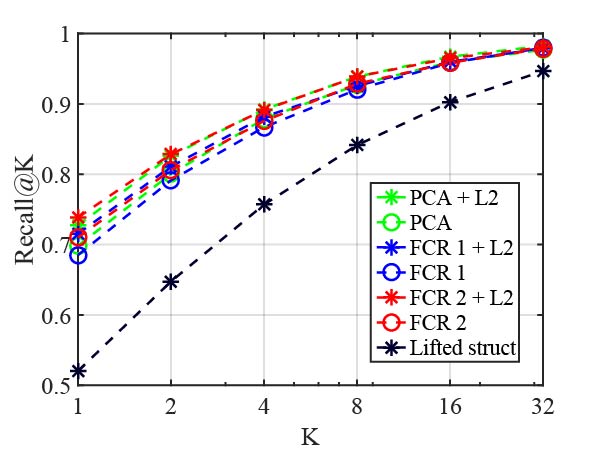}
		\end{minipage}		
		\caption{CAR: Comparisons between softmax-based features and lifted structured feature embedding \cite{song2016deep} on NMI (clustering), and Recall@K (retrieval) scores for the test set of the Stanford Cars 196 (CAR) dataset. The dimension of the feature used in the retrieval experiments is 64.}
		\label{fig:CAR}
	\end{minipage}
%\end{figure*}

%\begin{figure*}
	\begin{minipage}{0.90\linewidth}
%		\begin{minipage}{0.32\hsize}
             \hfill
		\begin{minipage}{0.45\hsize}
			\centering
			\includegraphics[width=\linewidth, bb=0 0 600 453]{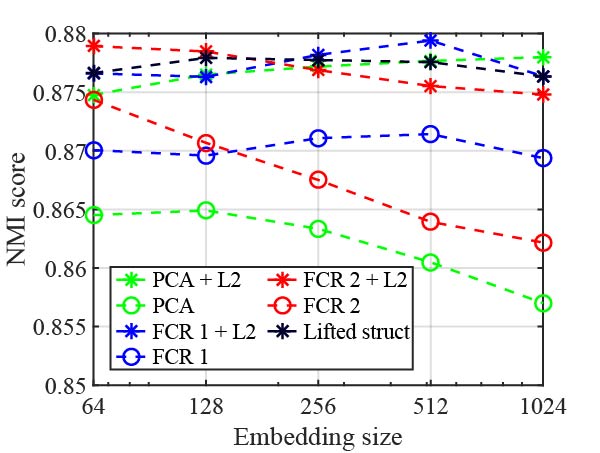}
		\end{minipage}
		\hfill
		\begin{minipage}{0.45\hsize}
			\centering
			\includegraphics[width=\linewidth, bb=0 0 600 453]{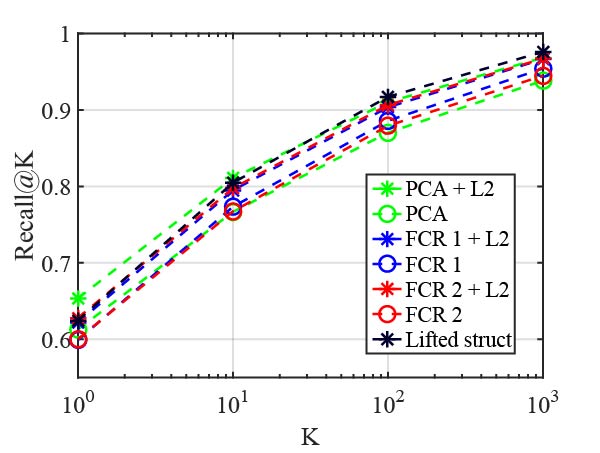}
		\end{minipage}		
		\caption{OP: Comparisons between softmax-based features and lifted structured feature embedding \cite{song2016deep} on NMI (clustering), and Recall@K (retrieval) scores for the test set of the Online Products (OP) dataset.The dimension of the feature used in the retrieval experiments is 64.}
		\label{fig:OP}
	\end{minipage}
\end{figure*}

%----------------------------------------------------
\section{Experiments}

In this section, we compared the deep features extracted from classification networks to those from state-of-the-art DML-based networks \cite{song2016deep,sohn2016improved,song2017learnable}.
The GoogLeNet architecture \cite{szegedy2015going} was used for all the methods---thus, the numbers of parameters are the same between DML-based networks and softmax-based features.
All the networks were fine-tuned from the weights pretrained on ImageNet \cite{russakovsky2015imagenet}.
We used the Caffe \cite{jia2014caffe} framework for the implementation.

\subsection{Comparisons between softmax-based features and DML-based features}
\label{sec:clustering}

Here, we give our evaluation of clustering and retrieval scores for the state-of-the-art DML methods \cite{song2016deep,sohn2016improved,song2017learnable} and for the softmax classification networks.
We used the Caltech UCSD Birds 200-2011 (CUB) dataset \cite{wah2011caltech}, the Stanford Cars 196 (CAR) dataset \cite{krause20133d}, and the Stanford Online Products (OP) dataset \cite{song2016deep}.
For CUB and CAR, we used the first half of the dataset classes for training and the rest for testing.
For OP, we used the training--testing class split provided.
The dataset properties are shown in \Tref{tbl:clustering_dataset}.
We emphasize that the class sets used for training and testing were completely different.

For clustering evaluation, we applied k-means clustering 100 times and calculated 
NMI (Normalized Mutual Information) \cite{manning2009introduction}; the value for $k$ was set to the number of classes in the test set.
For retrieval evaluation, we calculated Recall@K \cite{jegou2011product}.

In \Tref{tbl:CUB} and \Tref{tbl:CAR}, we show comparisons of the performance of clustering and retrieval using NMI and Recall@K scores, respectively, for CUB and CAR datasets.
We compared the softmax-based features, lifted structure\cite{song2016deep}, N-pair loss \cite{sohn2016improved} and the clustering loss \cite{song2017learnable}.
The results of the DML methods were quoted from the paper \cite{song2017learnable}. 
Regarding the lifted structure\cite{song2016deep}, the results in the parenthesis correspond to the scores we obtained from running the publicly available code ourselves, which we confirmed were almost the same as those in \cite{song2017learnable}. 
As we can see from \Tref{tbl:CUB} and \Tref{tbl:CAR}, softmax-based features outperformed DML features. 
The softmax-based features all performed well in the two datasets.

In OP dataset shown in \Tref{tbl:OP}, contrasting to CUB and CAR datasets, DML features outperform softmax-based features. We will make detailed analysis in the subsequent section.

%%%%%%%%%%%%%%%%%%%%%%%%%%%%%%

\begin{figure*}[t!]
	\begin{minipage}{0.90\linewidth}
\hfill
		\begin{minipage}{0.45\hsize}
			\centering
			\includegraphics[width=\linewidth, bb=0 0 600 453]{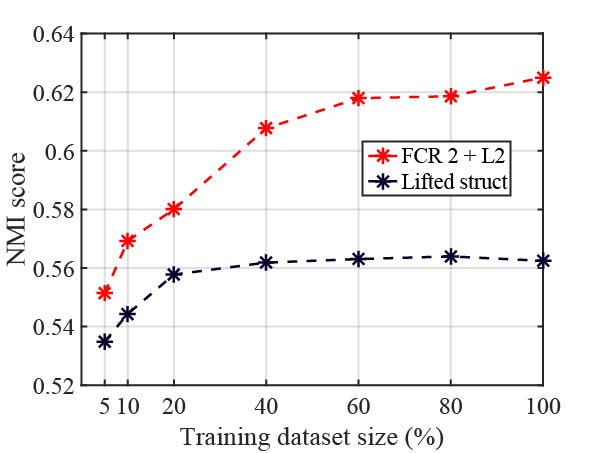}
		\end{minipage}
		\hfill
		\begin{minipage}{0.45\hsize}
			\centering
			\includegraphics[width=\linewidth, bb=0 0 600 453]{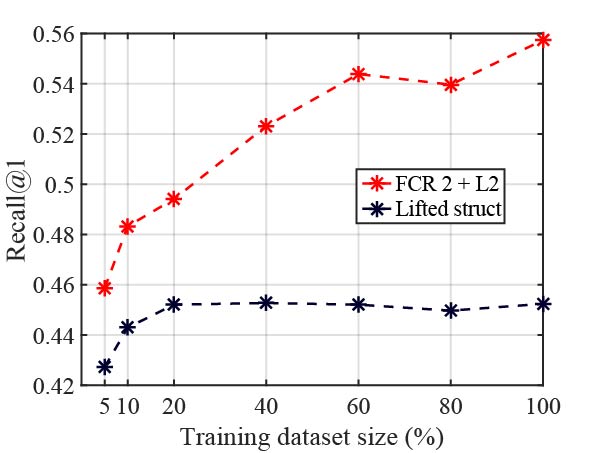}
		\end{minipage}	
		\caption{CUB: NMI (clustering), and Recall@K (retrieval) scores for test set of the Caltech UCSD Birds 200-2011 dataset under different dataset sizes. The feature dimensionality is fixed at 1024.}
		\label{fig:CUB_scale}
	\end{minipage}
	\\
	\begin{minipage}{0.90\linewidth}
\hfill
		\begin{minipage}{0.45\hsize}
			\centering
			\includegraphics[width=\linewidth, bb=0 0 600 453]{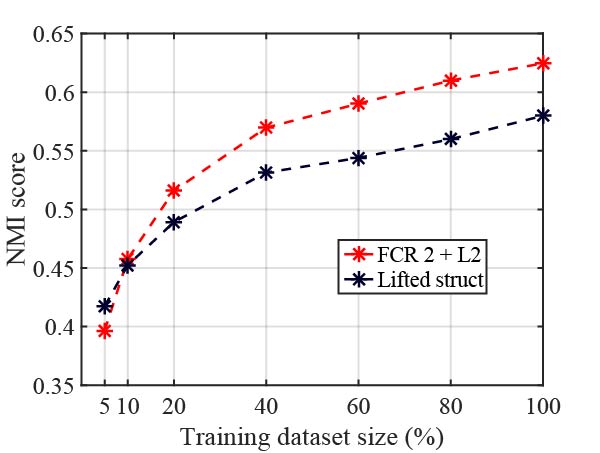}
		\end{minipage}
		\hfill
		\begin{minipage}{0.45\hsize}
			\centering
			\includegraphics[width=\linewidth, bb=0 0 600 453]{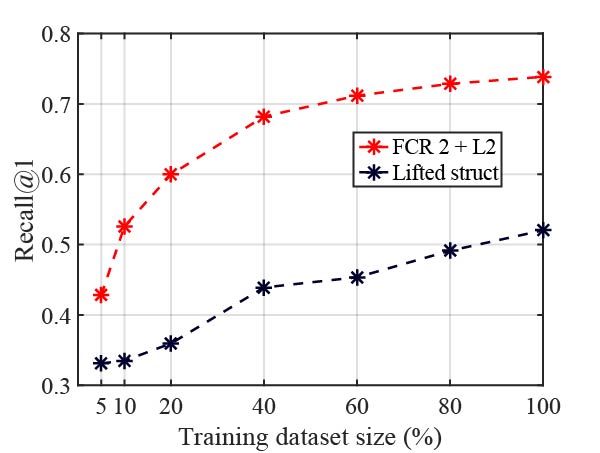}
		\end{minipage}	
		\caption{CAR: NMI (clustering), and Recall@K (retrieval) scores for test set of the Stanford Cars 196 dataset under different dataset sizes. The feature dimensionality is fixed at 256.}
		\label{fig:CAR_scale}
	\end{minipage}
\end{figure*}

\subsection{Detailed comparisons between softmax-based features and lifted structure embedding features}

We made detailed comparisons between softmax-based features and lifted structure embedding \cite{song2016deep} when changing dimensionalities and size of data. We conducted these experiments using the code available for lifted structure embedding \cite{song2016deep}.

Firstly, we show how the performance varies when changing the feature dimensionalities.
We changed the dimensionalities of softmax-based features via PCA, FCR1 and FCR2, and investigated how the performance of clustering and retrieval varied.
We compared them against those of lifted structure embedding of the same dimensionality.
 
For training, we multiplied the learning rates of the changed layers (output layers for all models and the fully connected layer added for FCR1 and FCR2) by 10.
The batch size was set to 128, and the maximum number of iterations for our training was set to 20,000, which was large enough for the three datasets to converge as mentioned in \cite{song2017learnable}.
These training strategies were exactly the same as those used in \cite{song2016deep}.

We show the results for CUB and CAR datasets in \Fref{fig:CUB} and in \Fref{fig:CAR}, respectively, under varying dimensionalities. The deep features extracted from the softmax-based classification networks outperformed the lifted structured feature embedding in clustering (NMI) and retrieval (Recall@K).

For clustering performance measured by NMI, all of the softmax models (PCA, FCR1, and FCR2) showed better scores than the lifted structured feature embedding. Regarding normalization, softmax-based features with L2 normalization showed better performance than those without normalization.
The NMI scores of PCA, FCR1 and FCR2 monotonically increased as the feature dimensionality increased for the CUB dataset (\Fref{fig:CUB}).
On the other hand, in CAR dataset (\Fref{fig:CAR}), the NMI scores of FCR2 and the lifted structure embeddings decreased from 256 dimensions and those of PCA and FCR1 were saturated above 256 dimensions.
This experimental result shows that 1024 dimensions is too large to represent the image classes of CAR dataset. 
It also implies that the feature dimensionality should be carefully considered in order to achieve best performance depending on the target data.

For retrieval performance measured by Recall@K metric, the softmax-based features also outperformed features of lifted structured feature embedding.
Regarding L2 normalization, features with normalization showed better score than without L2-normalization.

\Fref {fig:OP} shows the clustering and retrieval performance measured by NMI, and Recall@K, respectively, for the Online Products dataset.
Contrasting to CUB and CAR datasets, the softmax-based features with L2 normalization and the lifted structure embedding showed almost the same performance in the clustering and retrieval.
As shown in \Tref{tbl:clustering_dataset}, the OP dataset is very different from the CUB and CAR datasets in terms of the number of classes and the number of samples per class---the number of classes is 22k and the number of samples is 120k. The number of samples per class in the OP dataset is 5.3 on average, which is far smaller than the CUB and CAR dataset.

\subsection{The effect of the dataset scales}
From the results for these three datasets, we conjecture that the dataset size---that is the number of samples per class---has a considerable influence on softmax-based features.
Hence, we changed the size of datasets by sampling the images of CUB and CAR datasets for each class and ran the experiments again.
We constructed seven datasets of different sizes, containing 5, 10, 20, 40, 60, 80, and 100\% of the whole dataset, respectively.
Among them, 5\% corresponds to approximately 3 and 4 images per class in the CUB and the CAR dataset, respectively.
As shown in \Fref{fig:CUB_scale} and \Fref{fig:CAR_scale}, the differences between the scores for softmax and DML were small if the size of the training dataset was small.
The gap between softmax and DML became larger as the dataset size increased.
The softmax-based classifier was largely influenced by the size of the dataset.

\subsection{Distance metric with probability invariant shift}
\label{ssec:shift_vs_l2}
We define a distance metric that considers the softmax invariance
and null space of the linear projection matrix W. 
When two feature vectors are mapped to the same probability, 
the distance between the two becomes zero.
Assume a vector $\mathbf{u}$ such that
\begin{equation}
W\mathbf{u}=\mathbf{1}.
\end{equation}
The shift operation $\mathbf{x}+c\mathbf{u}$ has no influence on the softmax operation because $\mathrm{softmax}\left(W\mathbf{x}\right)=\mathrm{softmax}\left(W\left(\mathbf{x}+c\mathbf{u}\right)\right)=\mathrm{softmax}\left(W\mathbf{x}+c\mathbf{1}\right)$, where $c$ is an arbitrary constant.
$\mathbf{u}$ exists when the dimensionality of the feature $\mathbf{x}$ is larger than the number of classes to be classified.
$\mathbf{u}$ is represented by
\begin{equation}
\mathbf{u}=c_0W^\mathsf{T}\left(WW^\mathsf{T}\right)^{-1}\mathbf{1}+\sum_{i=1}^{D-1} c_i\mathbf{v}_i,
\end{equation}
where $W^\mathsf{T}\left(WW^\mathsf{T}\right)^{-1}$ is the pseudo-inverse of the linear projection matrix $W$, $\left\{\mathbf{v}_1\dots\mathbf{v}_{D-1}\right\}$ are the basis vectors that span the null space of $W$, and $\left\{c_0\dots c_{D-1}\right\}$ are arbitrary constants.
The shift operation is called the probability invariant shift in this paper.
Using the probability invariant shift, the distance between $\mathbf{x}_1$ and $\mathbf{x}_2$, defined below, removes
the effects of the softmax ambiguity and dimensionality reduction.
\begin{multline}
d\left(\mathbf{x}_1, \mathbf{x}_2\right)=\\
\min_{\left\{c_0,\dots,c_{D-1}\right\}} \left\lVert\mathbf{x}_1-\mathbf{x_2}-c_0W^\mathsf{T}\left(WW^\mathsf{T}\right)^{-1}\mathbf{1}-\sum_{i=1}^{D-1} c_i\mathbf{v}_i\right\rVert.
\end{multline}

In this section, we present comparative experiments on the distance with a probability invariant shift and with L2 normalization using the CUB and the CAR datasets.
Because the CUB and CAR have 100 and 98 classes in their training datasets respectively, we use $\{128, 256, 512, 1024\}$ dimensionality for the features for the experiments.

Tables \ref{tbl:shift_cub} and \ref{tbl:shift_car} show the results of the comparisons.
In all cases, the L2 normalization was the most effective.
The results demonstrated that the distance metric with a probability invariant shift had 
little effect on the clustering performance.

\begin{table}
	\centering
	\caption{NMI scores for the test set of the Caltech UCSD Birds 200-2011 (CUB) dataset: comparisons of the distance metric with a probability invariant shift and L2 normalization.}
	\label{tbl:shift_cub}
	\begin{tabular}{lcccc}
		\toprule
		&\multicolumn{4}{c}{dimensionality}\\\cmidrule(lr){2-5}
		&128&256&512&1024\\\midrule
		FCR 1&59.4&59.7&60.4&61.1\\
		FCR 1+Shift&59.2&59.1&59.3&59.3\\
		FCR 1+L2&\textbf{60.1}&\textbf{60.4}&\textbf{60.9}&\textbf{62.0}\\\midrule
		FCR 2&59.6&60.9&61.6&61.8\\
		FCR 2+Shift&59.3&60.0&60.0&60.1\\
		FCR 2+L2&\textbf{60.5}&\textbf{61.7}&\textbf{62.2}&\textbf{62.3}\\\bottomrule
	\end{tabular}
\end{table}
\begin{table}
	\centering
	\caption{NMI scores for the test set of the Stanford Cars 196 (CAR) dataset: comparisons of the distance metric 
with a probability invariant shift and L2 normalization.}
	\label{tbl:shift_car}
	\begin{tabular}{lcccc}
		\toprule
		&\multicolumn{4}{c}{dimensionality}\\\cmidrule(lr){2-5}
		&128&256&512&1024\\\midrule
		FCR 1&57.6&58.8&58.9&59.4\\
		FCR 1+Shift&57.4&58.2&57.9&57.8\\
		FCR 1+L2&\textbf{59.0}&\textbf{60.1}&\textbf{60.2}&\textbf{60.3}\\\midrule
		FCR 2&61.4&61.8&61.4&58.7\\
		FCR 2+Shift&61.2&61.3&60.7&58.2\\
		FCR 2+L2&\textbf{62.0}&\textbf{62.5}&\textbf{61.9}&\textbf{60.0}\\\bottomrule
	\end{tabular}
\end{table}

\section{Conclusion}
Because there was no equitable comparison in previous studies, we conducted comparisons of the softmax-based features and the state-of-the-art DML features using a design that would enable these methods to objectively demonstrate their true performance capabilities.
Our results showed that the features extracted from softmax-based classifiers performed better than those from state-of-the-art DML methods \cite{song2016deep,sohn2016improved,song2017learnable} on fine-grained classification, clustering, and retrieval tasks when the size of the training dataset (samples per class) is large.
The results also showed that the size of the dataset largely influenced the performance of softmax-based features. 
When the size of the dataset was small, DML showed better or competitive performance.
DML methods have advantages when the number of classes is very large and the softmax-based classifier is no longer applicable. 
In DML studies, softmax-based feature have rarely been compared fairly with DML-based feature under the same network architecture or with adequate fine-tuning. This paper revealed that the softmax-based features are still strong baselines. The results suggest that fine-tuned softmax-based features should be taken into account when evaluating the performance of deep features.

\subsection*{Limitations.}
\begin{itemize}
	\item When the number of classes is huge, it is hard to train classification networks due to GPU memory constraints.
DML-based methods are suitable for such cases because they do not need the output layer which is proportional to the number of classes.
	\item For cross-domain tasks, such as sketches to photos \cite{yu2016sketch,sangkloy2016sketchy} or aerial views to ground views \cite{lin2015learning}, DML is effective.
Classification-based learning needs complicated learning strategies like in \cite{catrejon2016learning}.
DML-based methods can learn cross-domain representation only by using a pair of networks.
	\item For datasets with continuous labels, DML-based methods might be helpful because classifier-based method cannot deal with them. However, most recent DML studies are specialized to datasets with discrete labels. To utilize the methods to datasets with continuous labels, some extensions are necessary.
\end{itemize}

%\bibliography{ref.bib}

\end{document}